\documentclass[a4paper]{article}

\usepackage{INTERSPEECH2021}
\usepackage{colortbl}

\title{Gender Bias and Universal Substitution Adversarial Attacks on Grammatical Error Correction Systems for Automated Assessment}
\name{Vyas Raina, Mark Gales}
\address{
  University of Cambridge}
\email{\{vr313, mjfg\}@cam.ac.uk}

\begin{document}

\maketitle
\begin{abstract}
  \let\thefootnote\relax\footnotetext{This work reports on research supported by Cambridge Assessment, University of Cambridge.}
  Grammatical Error Correction (GEC) systems perform a sequence-to-sequence task~\cite{DBLP:journals/corr/abs-1806-09030}, where an input word sequence containing grammatical errors, is corrected for these errors by the GEC system to output a grammatically correct word sequence. With the advent of deep learning methods, automated GEC systems have become increasingly popular. For example, GEC systems are often used on speech transcriptions of English learners as a form of assessment and feedback -  these powerful GEC systems can be used to automatically measure an aspect of a candidate's fluency. The count of \textit{edits} from a candidate's input sentence (or essay) to a GEC system's grammatically corrected output sentence is indicative of a candidate's language ability, where fewer edits suggest better fluency. The count of edits can thus be viewed as a \textit{fluency score} with zero implying perfect fluency. However, although deep learning based GEC systems are extremely powerful and accurate, they are susceptible to adversarial attacks: an adversary can introduce a small, specific change at the input of a system that causes a large, undesired change at the output~\cite{wang-zheng-2020-improving}. When considering the application of GEC systems to automated language assessment, the aim of an adversary could be to cheat by making a small change to a grammatically incorrect input sentence that conceals the errors from a GEC system, such that no edits are found and the candidate is unjustly awarded a perfect fluency score. \newline
  \indent Nevertheless most adversarial attack generation approaches in literature require multiple queries of the target system~\cite{DBLP:journals/corr/abs-1901-06796, raina-gales-2022-residue}. However, in the setting of language assessment, a candidate cannot query a GEC system. To overcome this issue, this work uses \textit{universal} adversarial attacks~\cite{Raina2020UniversalAA}, where the \textit{same} small change has to be made to any input sequence, such that the errors are concealed from the GEC system to obtain a perfect fluency score. As the candidates are non-native speakers of English, it is further required that the form of the attack has to be simple to apply.
     \begin{table}[htb!]
      \centering
      \begin{tabular}{lrrr}
      \toprule
           & FCE & BEA & CoNLL \\ \midrule
          F1 (\%) & 49.8 & 45.2 & 37.1 \\
          \bottomrule
      \end{tabular}
      \caption{GEC system performance}
      \label{tab:perf}
  \end{table}
  \vspace{-0.5cm}
  The simplest such attack is in the form of universal substitutions to exploit potential gender biases in a GEC system. For example, a candidate could replace all \textit{male} pronouns with \textit{female} pronouns, e.g, any occurrence of \textit{he} is replaced with \textit{she}. To determine the extent of threat of this form of adversarial attack, experiments were performed using a popular, publicly available Transformer-based GEC system, the \textit{Gramformer}~\cite{damodaran}, when applied to three benchmark GEC datasets~\cite{yannakoudakis-etal-2011-new, bryant-etal-2019-bea, ng-etal-2014-conll}, shown in Table \ref{tab:perf}.
   
  The impact of a universal gender pronoun substitution attack is shown in Table \ref{tab:gender}. For all datasets the GEC system is worryingly biased by the gender, where a candidate can reduce the number of edits made by the GEC system by simply swapping all male gender pronouns with female pronouns (\texttt{m2f}).
  
 \begin{table}[htb!]
     \centering
     \begin{tabular}{lrrrr}
     \toprule
         Substitution & FCE & BEA & CoNLL & \\ \midrule
         \texttt{m2f} & $-7.2$\% & $-2.8$\% & $-0.5$\% &$\downarrow$\\
         \texttt{f2m} & $+64.3$\% & $+15.3$\% &$+14.8$\% & $\uparrow$\\
         \bottomrule
     \end{tabular}
     \caption{Change (\%) in Avg. Edits with gender substitution.}
     \label{tab:gender}
 \end{table}
\vspace{-0.5cm}
  The gender pronoun substitution attack can be generalized to a universal substitution attack: a fixed dictionary mapping of word substitutions can be defined for some target words. When a target word appears in an input sequence it is replaced with its corresponding substitution word. For automated assessment with GEC, an adversary can learn and define the optimal dictionary of word mappings that when applied to any input deceives the GEC system into making no edits. The adversary can sell this dictionary to candidates looking to engage in mal-practice - this is a universal substitution attack approach that is agnostic to the original input sequence.\newline 
  \indent To mimic a realistic setting, the universal substitution dictionary is learnt using only the FCE train set and impact of the adversarial attack is evaluated on other test sets. For computational feasibility, the number of target words has to be limited, as identification of the optimal substitution word demands a greedy search through the English vocabulary. Selection of target words is thus hand-crafted: the most frequent words in the FCE train set, separately for each part of speech (POS), are identified. The universal learnt substituted words are matched in POS with the target words they replace. In this work, target words are restricted to nouns, adjectives or adverbs, e.g. it is found that the target noun \textit{life} should be substituted with the noun \textit{metamorphosis} to reduce number of edits. Table \ref{tab:attack} presents the impact of the universal substitution dictionary when applied to the unseen BEA and CoNLL test sets, where the dictionary has only a total of 14 target words (6 nouns, 4 adjectives, 2 adverbs and 3 gender pronouns). Note that results are presented only for the samples that are affected by the substitutions. It is interesting to note that even with such few target words there is a reduction in the number of edits made by the GEC system on unseen test sets.
\begin{table}[htb!]
    \centering
    \begin{tabular}{llr}
    \toprule
       Data &  No Attack & Sub Attack  \\ \midrule
        BEA & 2.665 & 2.512\\
        CoNLL & 2.554 & 2.437\\
        \bottomrule
    \end{tabular}
    \caption{Avg. number of GEC edits with Universal attack.}
    \label{tab:attack}
\end{table}

\end{abstract}

\bibliographystyle{IEEEtran}
\bibliography{mybib}

\newpage
\appendix
\newpage
~\newpage

\section{Substitution Search Words}
\renewcommand{\thesection}{\Alph{section}}
\renewcommand\thefigure{\thesection.\arabic{figure}} 
\setcounter{figure}{0}
\renewcommand\thetable{\thesection.\arabic{table}} 
\setcounter{table}{0}

Table \ref{tab:uni-sub-groups} enumerates the target (most frequent) search words from the FCE training set. These words were targeted for the universal substitutions for each part of speech (POS). The FCE training set sentences were used to greedily learn the substitution words for each target word, where the selected word is one that minimises GEC edits over the training set sentences. The FCE test set was used to identify the successful and unsuccessful universal substitutions, as given in colour-coded Tables: \ref{tab:uni-sub-JJ}, \ref{tab:uni-sub-CC}, \ref{tab:uni-sub-NN}, \ref{tab:uni-sub-PRPd} and \ref{tab:uni-sub-RB}.

\begin{table*}[htb!]
    \centering
    \begin{tabular}{l|l|l}
    \toprule
        Tag & Description & Words\\ \midrule
        CC & conjunction & $\underset{7368}{\text{and}}$, $\underset{1716}{\text{but}}$, $\underset{882}{\text{or}}$, $\underset{409}{\text{But}}$, $\underset{325}{\text{And}}$\\
        CD & numeral cardinal & $\underset{647}{\text{one}}$, $\underset{393}{\text{two}}$,\\
        DT & determiner & $\underset{12994}{\text{the}}$, $\underset{5209}{\text{a}}$, $\underset{1446}{\text{this}}$, $\underset{1001}{\text{some}}$, $\underset{986}{\text{The}}$, $\underset{983}{\text{all}}$, $\underset{719}{\text{that}}$, $\underset{560}{\text{an}}$,\\
        EX & existential there & $\underset{768}{\text{there}}$, $\underset{174}{\text{There}}$\\
        IN & preposition & $\underset{5045}{\text{in}}$, $\underset{5036}{\text{of}}$, $\underset{3200}{\text{for}}$, $\underset{2940}{\text{that}}$, $\underset{2054}{\text{because}}$, $\underset{1896}{\text{at}}$, $\underset{1705}{\text{with}}$, $\underset{1532}{\text{on}}$\\ 
        JJR & adjective comparative & $\underset{473}{\text{more}}$, $\underset{111}{\text{better}}$\\
        JJS & adjective superlative & $\underset{245}{\text{best}}$, $\underset{102}{\text{most}}$\\
        JJ & adjective & $\underset{788}{\text{good}}$, $\underset{573}{\text{other}}$, $\underset{391}{\text{last}}$, $\underset{356}{\text{different}}$, $\underset{355}{\text{many}}$, $\underset{349}{\text{great}}$, $\underset{343}{\text{much}}$, $\underset{332}{\text{new}}$\\
        MD & modal auxillary & $\underset{2162}{\text{would}}$, $\underset{1525}{\text{will}}$, $\underset{1347}{\text{can}}$, $\underset{974}{\text{could}}$, $\underset{553}{\text{should}}$, $\underset{217}{\text{must}}$, $\underset{203}{\text{ca}}$, $\underset{102}{\text{may}}$\\
        NNP & proper noun & $\underset{454}{\text{July}}$, $\underset{397}{\text{London}}$, $\underset{394}{\text{Pat}}$, $\underset{308}{\text{Danny}}$, $\underset{305}{\text{THE}}$, $\underset{298}{\text{Brook}}$, $\underset{257}{\text{First}}$, $\underset{223}{\text{TO}}$\\
        NNS & noun plural& $\underset{1110}{\text{people}}$, $\underset{584}{\text{clothes}}$, $\underset{423}{\text{things}}$, $\underset{371}{\text{activities}}$, $\underset{292}{\text{years}}$, $\underset{284}{\text{friends}}$, $\underset{284}{\text{students}}$, $\underset{243}{\text{discounts}}$\\
        NN & common noun& $\underset{1238}{\text{show}}$, $\underset{1137}{\text{time}}$, $\underset{840}{\text{money}}$, $\underset{730}{\text{life}}$, $\underset{516}{\text{school}}$, $\underset{477}{\text{advertisement}}$, $\underset{475}{\text{shopping}}$, $\underset{455}{\text{lot}}$\\
        PDT & pre-determiner& $\underset{447}{\text{all}}$\\
        POS & genitive marker & $\underset{503}{\text{'s}}$\\
        PRP &pronoun personal & $\underset{14490}{\text{I}}$, $\underset{4364}{\text{you}}$, $\underset{3875}{\text{it}}$, $\underset{2185}{\text{we}}$, $\underset{1940}{\text{me}}$, $\underset{1061}{\text{they}}$, $\underset{851}{\text{It}}$, $\underset{615}{\text{them}}$\\
        PRP\$ & pronoun possessive & $\underset{3006}{\text{my}}$, $\underset{1564}{\text{your}}$, $\underset{925}{\text{our}}$, $\underset{448}{\text{their}}$, $\underset{333}{\text{her}}$, $\underset{289}{\text{his}}$, $\underset{169}{\text{My}}$, $\underset{143}{\text{Your}}$\\
        RBR & adverb comparative & $\underset{426}{\text{more}}$\\
        RBS & adverb superlative & $\underset{255}{\text{most}}$,\\
        RB & adverb & $\underset{1852}{\text{n't}}$, $\underset{1774}{\text{not}}$, $\underset{1609}{\text{very}}$, $\underset{740}{\text{so}}$, $\underset{583}{\text{also}}$, $\underset{554}{\text{really}}$, $\underset{548}{\text{only}}$, $\underset{399}{\text{just}}$\\
        RP & particle & $\underset{347}{\text{up}}$, $\underset{301}{\text{out}}$\\
        VBD & verb past tense & $\underset{4027}{\text{was}}$, $\underset{1358}{\text{had}}$, $\underset{1026}{\text{were}}$, $\underset{485}{\text{did}}$, $\underset{390}{\text{went}}$, $\underset{378}{\text{started}}$, $\underset{304}{\text{said}}$, $\underset{244}{\text{told}}$\\
        VBG & verb present participle& $\underset{496}{\text{going}}$, $\underset{378}{\text{writing}}$, $\underset{239}{\text{looking}}$, $\underset{194}{\text{shopping}}$, $\underset{162}{\text{being}}$, $\underset{151}{\text{doing}}$, $\underset{128}{\text{evening}}$, $\underset{110}{\text{playing}}$\\
        VBN & verb past participle & $\underset{537}{\text{been}}$, $\underset{277}{\text{closed}}$, $\underset{143}{\text{seen}}$, $\underset{141}{\text{changed}}$, $\underset{122}{\text{written}}$, $\underset{103}{\text{done}}$\\
        VBP & verb present & $\underset{2252}{\text{have}}$, $\underset{1619}{\text{are}}$, $\underset{1329}{\text{am}}$, $\underset{805}{\text{do}}$, $\underset{760}{\text{think}}$, $\underset{593}{\text{'m}}$, $\underset{484}{\text{want}}$, $\underset{307}{\text{need}}$\\
        VB & verb & $\underset{2267}{\text{be}}$, $\underset{1165}{\text{like}}$, $\underset{974}{\text{have}}$, $\underset{700}{\text{go}}$, $\underset{627}{\text{do}}$, $\underset{527}{\text{know}}$, $\underset{463}{\text{see}}$, $\underset{425}{\text{take}}$\\
        VBZ & verb 3rd p singular & $\underset{3310}{\text{is}}$, $\underset{522}{\text{'s}}$, $\underset{428}{\text{has}}$, $\underset{104}{\text{does}}$\\
        WDT & WH-determiner & $\underset{890}{\text{which}}$, $\underset{382}{\text{that}}$\\
        WP & WH-pronoun& $\underset{687}{\text{what}}$, $\underset{438}{\text{who}}$, $\underset{127}{\text{What}}$\\
        WRB & WH-adverb& $\underset{871}{\text{when}}$, $\underset{542}{\text{how}}$, $\underset{281}{\text{When}}$, $\underset{218}{\text{where}}$, $\underset{215}{\text{why}}$\\
        \bottomrule
    \end{tabular}
    \caption{8 most common words by POS tag for FCE training set (Filtered to only contain grammatically incorrect sentences). Words with fewer than 100 occurrences are omitted.}
    \label{tab:uni-sub-groups}
\end{table*}

\begin{table*}[htb!]
    \tiny
    \centering
    \begin{tabular}{lllllllllllllllllllll}
    \toprule
        CC & CD & DT & IN & JJR & JJS & JJ & MD & NNS & NN & PRP & PRP\$ & RBR & RB & VBD & VBG & VBN & VBP & VB & VBZ & WP\\
        5 & 12 & 17 & 13 & 59 & 53 & 2300 & 12 & 8608 & 25955 & 16 & 7 & 4 & 1204 & 278 & 3166 & 3022 & 2 & 157 & 30 & 3\\
        \bottomrule
    \end{tabular}
    \caption{Vocab size for each POS}
    \label{tab:pos-vocab-size}
\end{table*}

\begin{table*}[htb!]
    \centering
    \begin{tabular}{lllrrrrrr}
    \toprule
       N  &  Orig & Sub & ALL & N1 & N2 & N4 & N5 & N6\\
       & & & \#2734 & \#58 & \#60& \#31 &\#57 &\#31\\\midrule
       0  & - & - &$\underset{\pm 1.752}{1.428}$ & $\underset{\pm 2.877}{2.069}$ & $\underset{\pm 2.081}{2.100}$ & $\underset{\pm 1.688}{2.129}$ & $\underset{\pm 2.237}{2.122}$ & $\underset{\pm 1.850}{1.903}$\\
       \cellcolor{green}1 & good & cavernous & $\underset{\pm 1.746}{1.426}$ & $\underset{\pm 2.704}{1.948}$ & \cellcolor{black}& \cellcolor{black} & \cellcolor{black} & \cellcolor{black}\\
       \cellcolor{red}2 & other & extraterrestrial &$\underset{\pm 1.751}{1.431}$ &\cellcolor{black} & $\underset{\pm 2.149}{2.300}$ & \cellcolor{black} & \cellcolor{black} & \cellcolor{black}\\
       \cellcolor{red}3 & last & last & $\underset{\pm 1.751}{1.431}$ & \cellcolor{black} & \cellcolor{black} & \cellcolor{black} & \cellcolor{black} & \cellcolor{black}\\
       \cellcolor{green}4 & different & dubious & $\underset{\pm 1.750}{1.429}$& \cellcolor{black} & \cellcolor{black} & $\underset{\pm 1.663}{2.032}$ & \cellcolor{black} & \cellcolor{black}\\
       \cellcolor{green}5 & many & dubious & $\underset{\pm 1.751}{1.424}$ & \cellcolor{black} & \cellcolor{black} & \cellcolor{black} & $\underset{\pm 2.214}{1.897}$ &\cellcolor{black}\\
       \cellcolor{green}6 & great & geopolitical & $\underset{\pm 1.748}{1.423}$ & \cellcolor{black}& \cellcolor{black} & \cellcolor{black} & \cellcolor{black} & $\underset{\pm 1.700}{1.710}$\\
       \cellcolor{red}7 & much & much & $\underset{\pm 1.748}{1.423}$& \cellcolor{black}& \cellcolor{black}& \cellcolor{black}& \cellcolor{black}& \cellcolor{black}\\
       \bottomrule
    \end{tabular}
    \caption{Universal substitution attack on finetuned Gramformer with $N$ most common \textbf{JJ} POS substituted. Results here on FCE test set. Average edits with N* filter data points that contain the target words substituted.}
    \label{tab:uni-sub-JJ}
\end{table*}

\begin{table*}[htb!]
    \centering
    \begin{tabular}{lllrr}
    \toprule
       N  &  Orig & Sub & Edits ALL\\
       & & & \#2734 samples\\\midrule
       0  & - & - & $1.428_{\pm 1.752}$ \\
       \cellcolor{red}1 & and & and & $1.428_{\pm 1.752}$  \\
       \cellcolor{red}\cellcolor{red}2 & but & but& $1.428_{\pm 1.752}$ \\
       \cellcolor{red}3 & or & or & $1.428_{\pm 1.752}$ \\
       \bottomrule
    \end{tabular}
    \caption{Universal substitution attack on finetuned Gramformer with $N$ most common \textbf{CC} POS substituted. Results here on FCE test set. Average edits with N* filter data points that contain the target words substituted.}
    \label{tab:uni-sub-CC}
\end{table*}

\begin{table*}[htb!]
    \centering
    \begin{tabular}{lllrrrrrrrr}
    \toprule
       N  &  Orig & Sub & ALL & N1 & N2 & N3 & N4 & N5 & N6 & N7\\
       & & & \#2734 & \#9 & \#107  & \#13 & \#85 & \#15 & \#0 & \#30\\\midrule
       0  & - & - & $\underset{\pm 1.752}{1.428}$ & $\underset{\pm 2.147}{1.889}$ & $\underset{\pm 1.609}{2.159}$ &$\underset{\pm 1.843}{2.308}$ & $\underset{\pm 2.098}{1.953}$ & $\underset{\pm 1.356}{1.867}$ & n/a & $\underset{\pm 2.246}{2.300}$\\
       \cellcolor{green}1 & show & trifecta & $\underset{\pm 1.752}{1.428}$&$\underset{\pm 2.224}{1.778}$ & \cellcolor{black}&\cellcolor{black}&\cellcolor{black} & \cellcolor{black} & \cellcolor{black}& \cellcolor{black}\\
       \cellcolor{green}2 & time & panama & $\underset{\pm 1.752}{1.421}$ &\cellcolor{black} & $\underset{\pm 1.662}{1.972}$&\cellcolor{black}&\cellcolor{black} & \cellcolor{black} & \cellcolor{black}& \cellcolor{black}\\
       \cellcolor{green}3 & money & topsoil & $\underset{\pm 1.754}{1.421}$ &\cellcolor{black}&\cellcolor{black}&$\underset{\pm 2.088}{2.231}$ &\cellcolor{black} & \cellcolor{black} & \cellcolor{black}& \cellcolor{black}\\
       \cellcolor{green}4 & life & metamorphosis & $\underset{\pm 1.755}{1.416}$ & \cellcolor{black} & \cellcolor{black} &\cellcolor{black}& $\underset{\pm 2.178}{1.824}$&\cellcolor{black}&\cellcolor{black} & \cellcolor{black}\\
       \cellcolor{green}5 & school & trifecta & $\underset{\pm 1.749}{1.411}$ &\cellcolor{black} & \cellcolor{black} & \cellcolor{black} &\cellcolor{black} & $\underset{\pm 1.265}{1.200}$ & \cellcolor{black}& \cellcolor{black}\\
       \cellcolor{red}6 & advertisement & kpn& $\underset{\pm 1.749}{1.411}$ & \cellcolor{black} & \cellcolor{black} & \cellcolor{black} & \cellcolor{black} & \cellcolor{black} &n/a& \cellcolor{black}\\
       \cellcolor{green}7 & shopping & bridgette & $\underset{\pm 1.756}{1.410}$ & \cellcolor{black}& \cellcolor{black}& \cellcolor{black}& \cellcolor{black}& \cellcolor{black}& \cellcolor{black} & $\underset{\pm 2.189}{2.200}$\\
       \bottomrule
    \end{tabular}
    \caption{Universal substitution attack on finetuned Gramformer with $N$ most common \textbf{NN} POS substituted. Results here on FCE test set. Average edits with N* filter data points that contain the target words substituted.}
    \label{tab:uni-sub-NN}
\end{table*}

\begin{table*}[htb!]
    \centering
    \begin{tabular}{lllrrr}
    \toprule
       N  &  Orig & Sub & Edits ALL & Edits N6\\
       & & & \#2734 samples & \# 9 samples\\\midrule
       0  & - & - & $1.428_{\pm 1.752}$ & $2.778_{\pm 3.420}$\\
       \cellcolor{red}1 & my & my &  $1.428_{\pm 1.752}$ & \cellcolor{black}\\
       \cellcolor{red}2 & your & your & $1.428_{\pm 1.752}$ & \cellcolor{black}\\
       \cellcolor{red}3 & our & our & $1.428_{\pm 1.752}$& \cellcolor{black}\\
       \cellcolor{red}4 & their & their & $1.428_{\pm 1.752}$& \cellcolor{black}\\
       \cellcolor{red}5 & her & her & $1.428_{\pm 1.752}$& \cellcolor{black}\\
       \cellcolor{green}6 & his & my & $1.427_{\pm 1.754}$ & $2.667_{\pm 3.391}$\\
       \bottomrule
    \end{tabular}
    \caption{Universal substitution attack on finetuned Gramformer with $N$ most common \textbf{PRP\$} POS substituted. Results here on FCE test set. Average edits with N* filter data points that contain the target words substituted.}
    \label{tab:uni-sub-PRPd}
\end{table*}

\begin{table*}[htb!]
    \centering
    \begin{tabular}{lllrrrrr}
    \toprule
       N  &  Orig & Sub & Edits ALL & Edits N1& Edits N3 & Edits N4 & Edits N6\\
       & & & \#2734 samples & \#181 samples & \#75 samples & \#28 samples & \#69 samples\\\midrule
       0  & - & - & $1.428_{\pm 1.752}$ & $2.061_{\pm 2.317}$ & $2.027_{\pm 1.708}$ & $2.107_{\pm 1.618}$ & $1.783_{\pm 1.688}$\\
       \cellcolor{red}1 & very & stylistically & $1.432_{\pm 1.761}$& $2.122_{\pm 2.401}$& \cellcolor{black} & \cellcolor{black}& \cellcolor{black}\\
       \cellcolor{red}2 & so & so & $1.432_{\pm 1.761}$ & \cellcolor{black}& \cellcolor{black}& \cellcolor{black} & \cellcolor{black}\\
       \cellcolor{green}3 & also & noticeably & $1.431_{\pm 1.760}$ & \cellcolor{black} & $1.947_{\pm 1.731}$& \cellcolor{black}& \cellcolor{black}\\
       \cellcolor{green}4 & really & romantically & $1.430_{\pm 1.761}$ &\cellcolor{black} & \cellcolor{black} & $2.036_{\pm 1.753}$& \cellcolor{black}\\
       \cellcolor{red}5 & only & only &  $1.430_{\pm 1.761}$ &\cellcolor{black} & \cellcolor{black} & \cellcolor{black}& \cellcolor{black}\\
       \cellcolor{red}6 & just & passionately & $1.430_{\pm 1.766}$& \cellcolor{black}& \cellcolor{black}& \cellcolor{black} & $1.797_{\pm 1.820}$\\
       \bottomrule
    \end{tabular}
    \caption{Universal substitution attack on finetuned Gramformer with $N$ most common \textbf{RB} POS substituted. Results here on FCE test set. Average edits with N* filter data points that contain the target words substituted.}
    \label{tab:uni-sub-RB}
\end{table*}

\begin{table*}[htb!]
    \centering
    \begin{tabular}{lp{5.5cm}rrr}
    \toprule
        Attack & Sub & Edits ALL & Edits (m) & Edits (f)\\
        & & \#2734 & \#40 & \#110 \\\midrule\midrule
        None & None & $1.428_{\pm 1.752}$ & $2.100_{\pm 2.262}$ & $0.564_{\pm 1.080}$\\
        \cellcolor{green}m2f & \{his:her, him:her, he:she, Mr:Mrs\} & $1.425_{\pm 1.751}$ & $1.950_{\pm 2.087}$& \cellcolor{black}\\
        \cellcolor{red}f2m & \{her:his, hers:his, she:he, Mrs:Mr\} &$1.442_{\pm 1.751}$ & \cellcolor{black}&$0.927_{\pm 1.139}$\\
        \bottomrule
    \end{tabular}
    \caption{FCE: Universal gender substitution attack}
    \label{tab:sub-gender}
\end{table*}

\end{document}